\documentclass{article}

\PassOptionsToPackage{numbers, sort, square}{natbib}

\usepackage[preprint]{neurips_data_2024}

\usepackage{amsmath}
\usepackage[utf8]{inputenc} 
\usepackage[T1]{fontenc}    
\usepackage{hyperref}       
\usepackage{url}            
\usepackage{booktabs}       
\usepackage{amsfonts}       
\usepackage{nicefrac}       
\usepackage{microtype}      
\usepackage{xcolor}         
\usepackage{graphicx}

\usepackage{dirtree}
\usepackage{float}  
\usepackage{color} 
\usepackage{listings} 
\usepackage{courier} 
  
\title{Human-centered In-building Embodied Delivery Benchmark}

\vspace{-10pt}
\author{
Zhuoqun Xu \textsuperscript{1} $^{*}$\\
  \And
  Yang Liu \textsuperscript{2}\\
  \And
  Xiaoqi Li \textsuperscript{3}\\
  \And
  Jiyao Zhang \textsuperscript{3}\\
  \And
  Hao Dong \textsuperscript{3} $^{\dagger}$
  \AND
  \vspace{-25pt}
  \\
  \textsuperscript{1} Polar Research General Services,
  \textsuperscript{2} Samsung R\&D Institute China - Beijing,
  \textsuperscript{3}Peking University \\
  \vspace{9pt}
  $^{*}$ zhuoqun.xu@prgsorg.com, $^{\dagger}$ hao.dong@pku.edu.cn
}
  
\begin{document}

\maketitle
\vspace{-15pt}
\begin{abstract}
Recently, the concept of embodied intelligence has been widely accepted and popularized, leading people to naturally consider the potential for commercialization in this field. In this work, we propose a specific commercial scenario simulation --- human-centered in-building embodied delivery. Furthermore, for this scenario, we have developed a brand-new virtual environment system from scratch, constructing a multi-level connected building space modeled after a polar research station. This environment also includes autonomous human characters and robots with grasping and mobility capabilities, as well as a large number of interactive items. Based on this environment, we have built a delivery dataset containing 13k language instructions to guide robots in providing services. We simulate human behavior through human characters and sample their various needs in daily life. Finally, we proposed a method centered around a large multimodal model to serve as the baseline system for this dataset. Compared to past embodied data work, our work focuses on a virtual environment centered around human-robot interaction for commercial scenarios. We believe this will bring new perspectives and exploration angles to the embodied community. Our work has been hosted in the CVPR 2024 Embodied Workshop\footnote{\url{https://embodied-ai.org/}} (PRS Challenge\footnote{\url{https://prsorg.github.io/}}). 
\vspace{5pt}

\end{abstract}
\section{Introduction}
\label{Introduction}
With the rapid development of embodied robotic technology, people are gradually becoming aware of its tremendous potential in various fields. Concurrently, there has been a surge of discussions and explorations within the community regarding embodied skill scenarios, such as navigation \cite{wang2024find, hao2020towards}, manipulation \cite{li2023manipllm, jin2024robotgpt}, and instruction following \cite{brohan2022can, yenamandra2023homerobot}, leading to the proposal of a series of models. 
Although the skill scenarios are diverse, people are concerned that the current skill scenarios are designed to be overly simplistic for commercial application scenarios \cite{fu2024mobile}. And there exists a noticeable gap between skill scenarios and commercial application scenarios. Specifically, it is widely believed that existing skill scenarios may be inadequate in fully reflecting the potential issues encountered in actual commercial environments and do not accurately capture users' more precise interaction needs with embodied robots \cite{bousmalis2023robocat}. Therefore, we argue that this inconsistency with real-world commercial scenarios has hindered the emergence of novel topics within the embodied AI community in recent years. Therefore, we suggest that exploring scenarios closer to real-world commercial applications can help further the development of the embodied AI community \cite{zador2023catalyzing, yang2023learning}.

\begin{figure}[t]
  \centering
  \includegraphics[width=\linewidth]{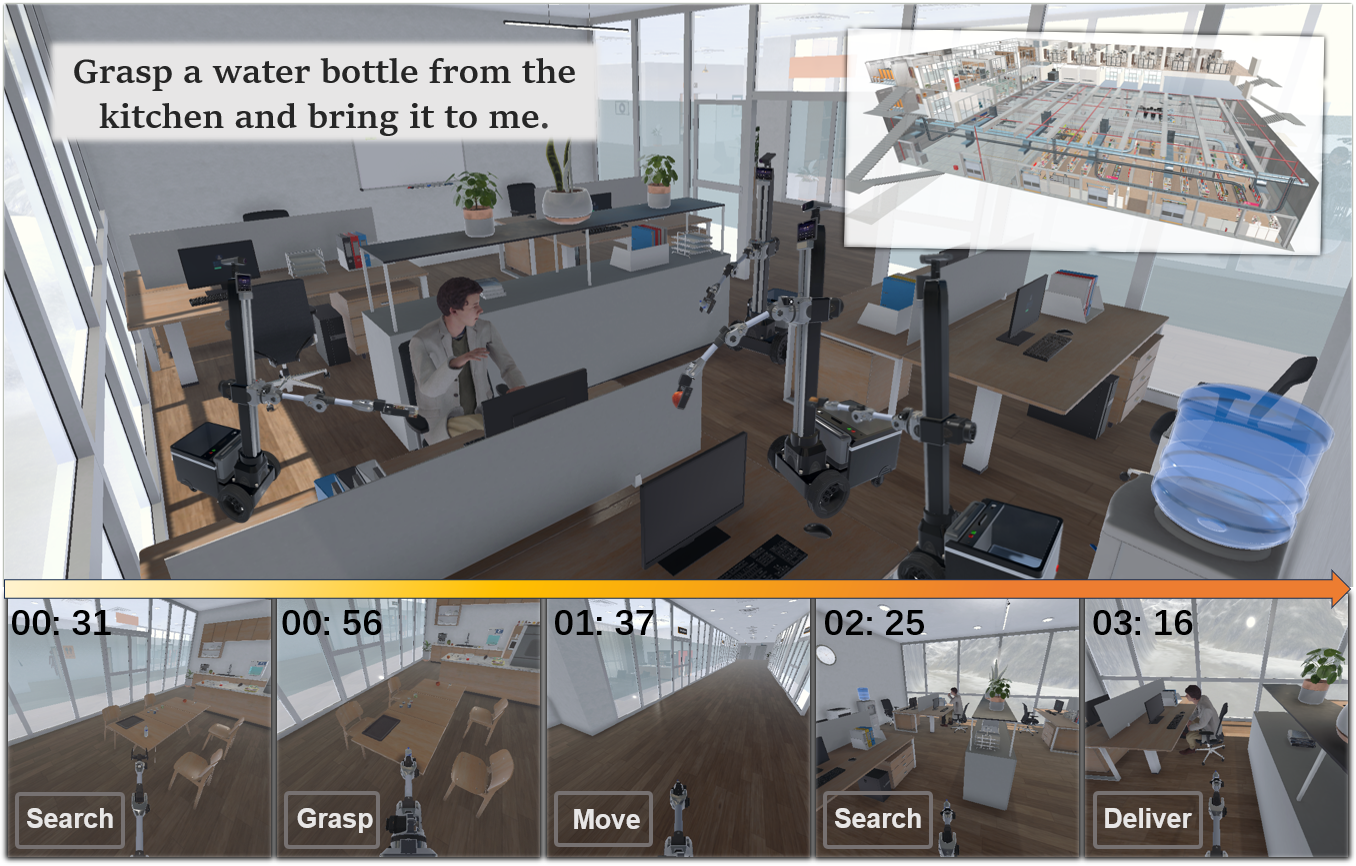}
  \vspace{-15pt}
  \caption{Human-centered in-building embodied delivery describes a task that originates from a real commercial delivery scenario. It mainly refers to the precise delivery service for users in private spaces where external delivery services cannot be used, achieved through embodied robots. This task typically requires the robot to locate the target item based on the user's needs (e.g. \textit{grasp a water bottle from the kitchen and bring it to me.}) across multiple rooms within the three-story building (a polar research station building, \textit{See the thumbnail in the top right corner}) and ultimately deliver it to the designated location/person. The robot needs to consider the user's context (behavior or schedule), as the user will be moving around the building according to their own goals during the delivery.}
  \vspace{-15pt}
  \label{overview}
\end{figure}
In this work, we focus on simulating and data construction for a highly anticipated express delivery service scenario called human-centered in-building delivery. In today's society, precise and efficient delivery services are crucial to the success of many top companies (such as Amazon, JD, and Meituan). However, unlike large-scale transshipment centers and external express delivery services that rely on public transportation, the last step of the delivery stage faces significant challenges. For instance, private spaces like company in buildings or high-security residential areas often prohibit external delivery services due to various security and management considerations. Moreover, people typically move around inside buildings to meet their needs and purposes. This delivery process can impose tangible pressure on customers. Therefore, precise item delivery to specified persons in private spaces represents a significant opportunity for robotic services. In order to explore this scenario, our contributions can be divided into the following parts:

\textbf{Scenario \& Task Definition.} The commercial scenario is characterized by the need to account for numerous complex and intervening factors. We analyze the scenario and pinpoint the critical elements of the delivery service. Followed by formulating task objectives and definitions, which includes establishing the task's premises, context, framework, and scope (Section \ref{scenario}).

\textbf{Simulation Environment.} Grounded in the task setting and business requirements, we have constructed from scratch a novel simulation environment modeled after a real-world polar research station (referred to as the Polar Research Station Environment, PRS). This environment comprises a three-story building interconnected by stairs and a functional elevator. It integrates common human societal scenarios into a community-like pattern, such as bedrooms, gyms, offices, laboratories, medical rooms, wards, living rooms, leisure spaces, etc. This design aims to cover as a wide range of everyday scenarios within the building as possible. Additionally, to simulate daily activities for delivery services, the environment includes over a dozen virtual human characters engaging in activities according to their individual intentions. Furthermore, we provide a range of interactive objects to support the tasks. Lastly, we have designed a robotic simulation with grasping and moving capabilities to serve human character agents (Section \ref{Simulation}).

\textbf{Dataset.}  In constructing delivery service data, we initially utilize the large language model (LLM) to generate reasonable daily activities and varied demands for virtual characters based on their profiles. Consequently, the robot is required to locate and deliver the appropriate objects to meet the human characters' demands and accomplish task objectives. We continually generate diverse data by modifying character needs, daily routines, and target objects. Furthermore, we incorporate a manual review and refinement stage to ensure the balance of the task data (Section \ref{Dataset}).

\textbf{Baseline.} We propose an LMM-based approach as the baseline method, employing a modular architecture encompassing language instruction analysis, multimodal target search, and robotic action execution (Section \ref{Baseline}). 

Therefore, we will gradually introduce these contents. Due to our substantial workload, additional content will be included in the supplementary materials.


\section{Background}
\label{Related Work}

In recent years, the concept of embodied AI has been widely recognized and popularized \cite{das2018embodied, gan2021threedworld, brohan2022rt, brohan2023rt}. People have been actively exploring the capabilities of this new form of intelligent entity \cite{wu2024unigarmentmanip}, such as embodied instruction following \cite{ pashevich2021episodic, li2023vision, padmakumar2022teach}, visual navigation \cite{hao2020towards}, and manipulation grasping \cite{wu2024learning}. Furthermore, with the introduction of large models, significant progress has also been made in the field \cite{mousavian20196, murali20206}. However, while researchers, investors, and engineers generally believe that existing skill-driven scenarios may demonstrate the potential of embodied robots \cite{jiang2022vima, gao2022dialfred}, their performance in comprehensive business scenarios remains uncertain, leading to widespread concern.

To mitigate this issue, we believe that introducing simulations of commercial scenarios might be a potential solution. The main difference from current skill-learning-oriented scenarios \cite{padalkar2023open, mandlekar2023mimicgen} is that commercial scenarios typically prioritize meeting human needs. This not only requires the integration of multiple skills \cite{wu2023learning} to achieve service objectives but also entails incorporating elements such as human-robot interaction \cite{long2023discuss}, scenario diversity \cite{deitke2020robothor}, and human behavior portrayal. The benefits of doing so are twofold: firstly, it can make robot training more closely resemble real commercial scenarios, and secondly, it can introduce new, more specific topics to the community, further promoting the community's evolution towards commercialization.

Furthermore, we systematically examined a large number of existing virtual environment systems (such as AI2thor\cite{kolve2017ai2}, Habitat \cite{puig2023habitat}, BEHAVIOR-1k \cite{li2023behavior}, etc.\cite{makoviychuk2021isaac, handa2023dextreme, james2020rlbench}), which generally struggle to simultaneously support the depiction of commercial scenarios requiring interconnected multi-level architectural spaces, diverse and multi-functional social spaces (such as laboratories, medical rooms), customizable interactive human character and behavior, a plethora of interactable items, and continuously changing motion states supported by physics engines \cite{todorov2012mujoco, haviland2023manipulator, haviland2023manipulator2}. Thus, we constructed the aforementioned simulation environment from scratch which is inspired by the polar research stations from real world . Thus, We chose the human-centered in-building delivery service as an initial exploration into simulating embodied commercial scenarios.

\section{Scenario Analysis \& Task Definition}
\label{scenario}

First, we need to analyze and abstract the commercial scenario in order to generate actionable tasks. In the context of precise in-building delivery services, we have identified several potential key factors:
\begin{itemize}
\item Robots operate within a relatively fixed building space.
\item  The residents within the building are the recipients of the service, and they typically move throughout the building based on personal needs and objectives. Robots can access relevant information about the recipients to better locate and identify them.
\item  The transportation service may cover a substantial area, involving different floors and rooms.
\item  Robots typically need to understand human instructions in order to search for and retrieve the correct target items, and deliver them to the designated recipients.
\end{itemize}
Based on the aforementioned scenario requirements, we provide the following task definition and settings, as shown in the Table \ref{table-task-setting}.

\begin{table}[ht]
    \centering
    \caption{Human-centered in-building embodied delivery task setting.}
    \label{table-task-setting}
    \begin{tabular}{|p{3cm}|p{10cm}|}
        \hline
        \textbf{Task Setting} & \textbf{Content} \\
        \hline
        Purpose & Deliver the requested item to the vicinity of the designated character. \\
        \hline
        Delivery Items & Items in the environment that can be grabbed and moved. (see item example in Figure \ref{task-information}) \\
        \hline
        Customers & Ten virtual human characters with different daily activities inside the building. They will move within the building for their own purposes. \\
        \hline
        
        Spatial Scope & The reachable areas within different rooms of a three-story building. \\
        \hline
        Time Setting & Real-world time, but simulation can be accelerated. \\
        \hline
        Customer Description & Self introduction and personal image photos, such as "\textit{I'm John, a supervisor who's often busy with meetings and office work. ... my office is  Room 2 on second floor. ... with a middle-aged man in a white shirt ... that's me.}" (see the personal image in Figure \ref{task-information}).
        \\
        \hline
        Scenario Map & 2D projected obstacle map of scenario, and pre-sampled panoramic photos at various locations on the map (see panoramic image of sampling position in 2D obstacle map in Figure \ref{task-information}). \\
        \hline
        Robot Positioning & We adopt relative localization rules for robot positioning, where its initial position is always set to (0, 0, 0). \\
        \hline
        Robot Actions & Movement, joint control, and manipulation. \\
        \hline
        Robot Skills & Local navigation by coordinate, 6-DOF visual grasping, and pose adjustment. \\
        \hline
        Sensors & Two RGB-D cameras (head and arm), tactile sensors. \\
        \hline
        Customer Instruction & Describe the goal, Identify the target object, describe its location, and confirm the target person of the delivery. For example, ``\textit{Fetch the blue-packaged \textbf{water bottle} from the wooden \textbf{dining table} in the kitchen and deliver it to \textbf{Imani}, the woman in the blue shirt with black glasses, in the kitchen room}''. \\
        \hline
        Success Criteria & Place the target object within a 3 meter range of the target person. \\
        \hline
        Constraints & Completion within 8 minutes without any dangerous collisions and unavailability of environmental metadata. \\
        \hline
    \end{tabular}
\end{table}

\begin{figure}[h]
  \centering
  \includegraphics[width=0.7\linewidth]{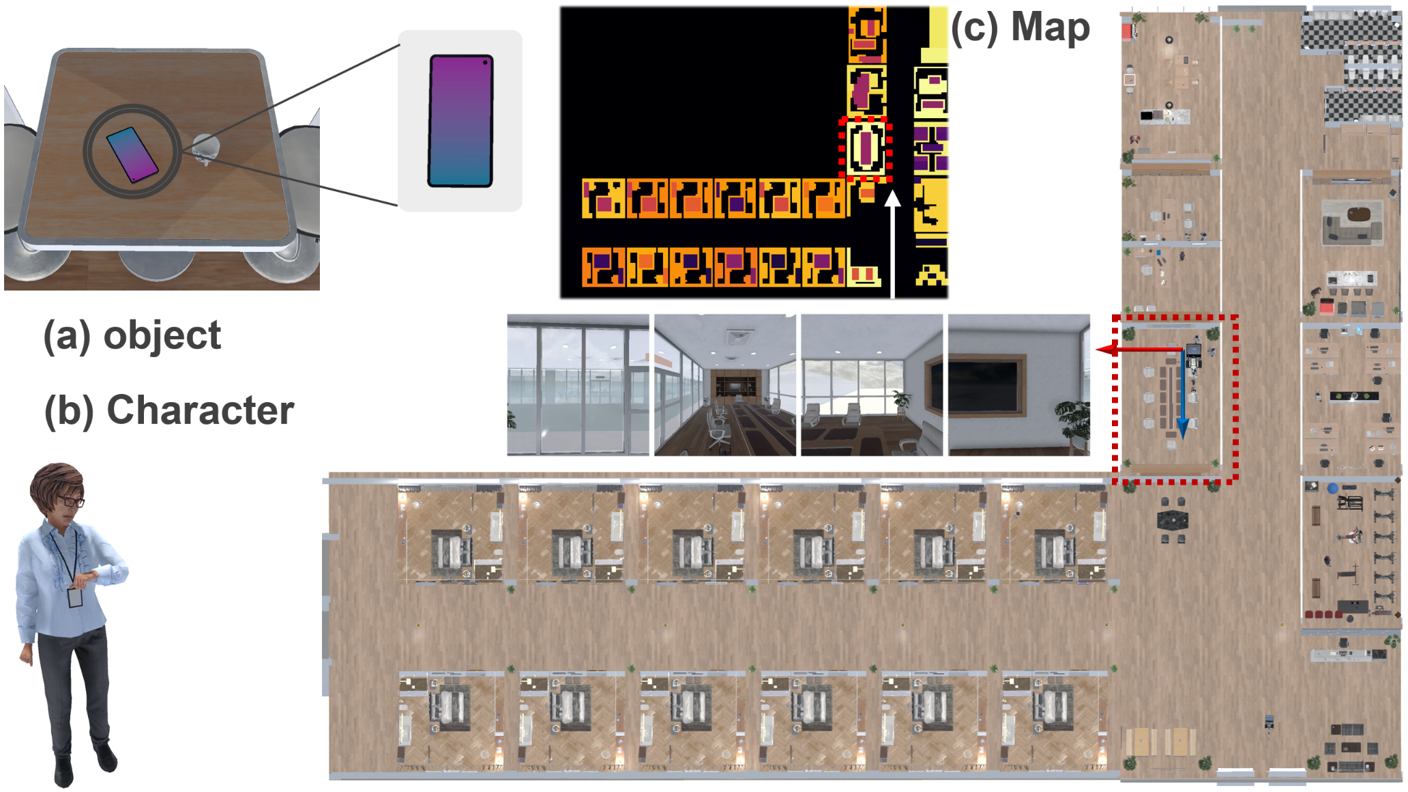}
  \caption{The available information in task.}
  \label{task-information}
\end{figure}

The distinction in design between skill scenarios and commercial scenarios lies in their objectives. Skill scenarios tend to focus on exploring the efficiency of a particular skill under given conditions. On the other hand, commercial scenarios exploration primarily revolves around identifying which conditions and information are most effective for achieving the ultimate goal in that scenario. Therefore, for this delivery task, we strive to provide as comprehensive and diverse information as possible to assist the robot in completing the task. Additionally, we continuously optimize the scenario and task design based on feedback from dataset users by adding more information channels.

\section{Simulation Environment}
\label{Simulation}

\begin{figure}[h]
  \centering
  \includegraphics[width=\linewidth]{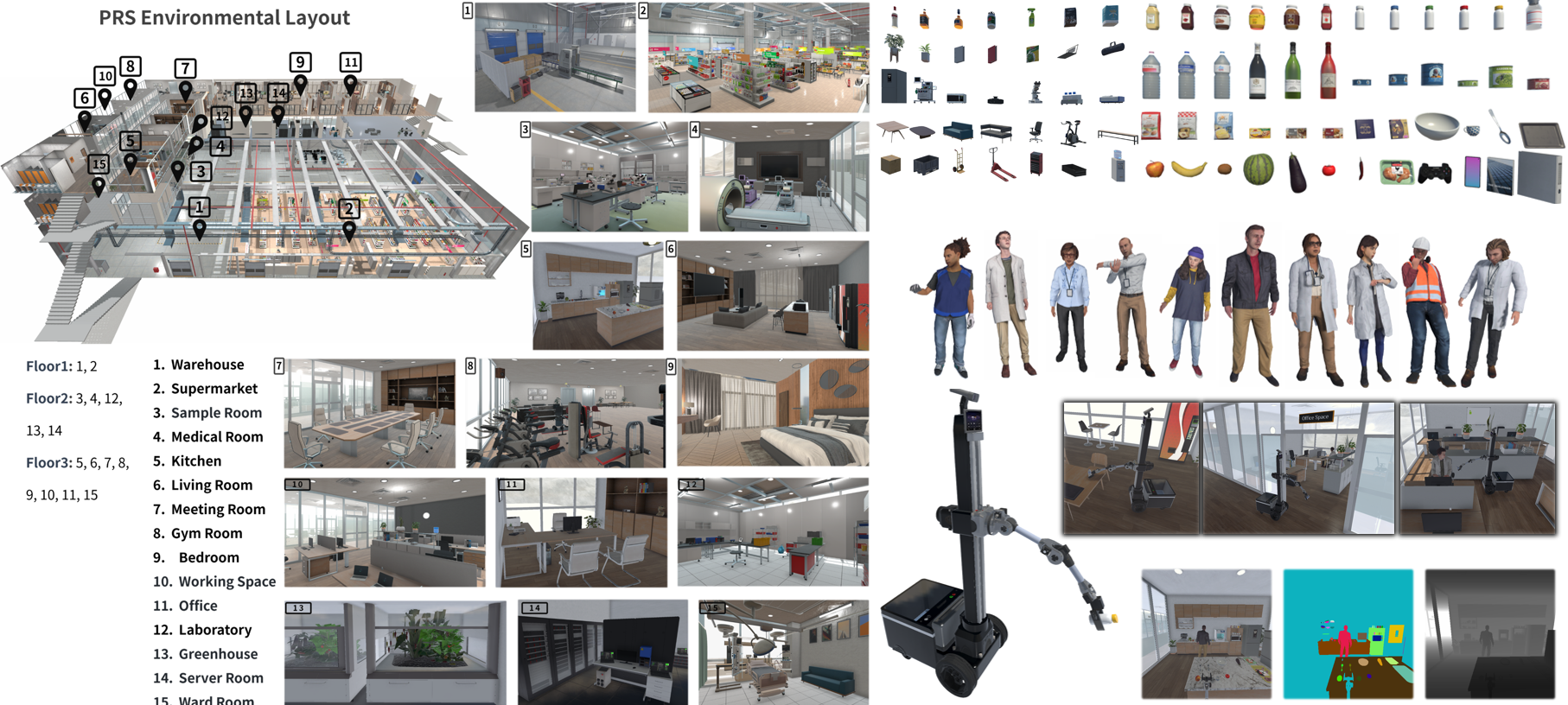}
  \caption{Environment includes three-story buildings, items, human characters, and robot.}
  \label{prs-env}
\end{figure}

Virtual environments typically need to meet the task requirements. Clearly, to depict corresponding commercial scenarios, existing environments are still constrained by factors such as the richness of the scene, the complexity of space, character portrayal, continuous environmental state systems, long-term operation, and the setup of items and robots. Therefore, we construct a brand-new virtual environment to support the tasks, as shown in Figure \ref{prs-env}. Next, we will introduce the main features of the virtual environment.

\textbf{Social Scenarios \& Space with 
Height}. As we mentioned, we need diverse common social scenarios. In existing work, we often see common indoor spaces such as kitchens, bedrooms, and living rooms, but less common are places like supermarkets, medical rooms, and studios. However, activities in different places vary greatly, and the premise of depicting diverse human activities in commercial scenarios is to include these settings. Moreover, it is the people constantly moving within these spaces that give them unique semantics. Additionally, we notice that existing work often confines scenes to a "flat" plane, with rare descriptions of "space with height", greatly limiting the spatial utilization of virtual environments in depicting complex scenes. Our virtual environment takes these factors into consideration.

\textbf{Human Character}. As mentioned earlier, the behavior of humans in commercial scenarios needs emphasis. The activities of the robot actually revolve around human activities. Therefore, in our virtual environment, we support a human character system that controls goals, actions, and interactions. In this task, we mainly adopt various forms of daily activities (working, resting, simple socializing, etc.) to depict the actions of characters. Since the delivery task is closely related to the positions of characters, we primarily drive the movement of characters within the building based on schedule information.

\textbf{Continuous Environment State}. Our virtual environment is primarily driven by a physics engine at its core, containing items with physical properties, so almost all movements are continuous (with exceptions for specific object state changes controlled by scripts and interfaces). Even when we use robot control interfaces similar to the ALFRED style \cite{shridhar2020alfred}(AI2-THOR\cite{kolve2017ai2}), such as ''\textit{pick\_obj()}'', the movements it executes require real-time implementation through continuous body control.

\textbf{Robot Configuration}. We use a robot with grasping and movement capabilities. It is equipped with visual perception (RGB-D) and simple tactile perception based on rigid body collision. At the core, we have prepared various control methods for it. Users can control the robot either through an ALFRED-style interface (typically invoked by high-level action and LMMs with object segmentation) or through a ROS-like interface.

More information can be found on the webpage \footnote{\url{https://prsorg.github.io/}}.

\begin{table}[h] 
\centering
\setlength{\tabcolsep}{1pt}
\caption{Comparison between PRS delivery tasks and existing dataset.}
\label{table-compare}
\begin{tabular}{c|ccccccc} 
\toprule 
{\small Benchmark}& {\small PRS}&  {\footnotesize ALFRED\cite{shridhar2020alfred}} &{\footnotesize EQA\cite{das2018embodied}} &{\footnotesize VirtualHome\cite{puig2018virtualhome}}&{\footnotesize BEHAVIOR-1K\cite{li2024behavior}}&{\footnotesize Habitat\cite{puig2023habitat}}&{\footnotesize iGibson\cite{shen2021igibson}}
\\  
\midrule
{\footnotesize Directive} &\checkmark & \checkmark & \checkmark & - & \checkmark  & \checkmark & - \\ 
{\footnotesize Continuous State} &\checkmark & -& - & - & \checkmark  & \checkmark & \checkmark\\   
{\footnotesize Articulated Joints} &\checkmark& - & - & - & \checkmark &- & -\\ 
{\footnotesize Mobile Characters} &\checkmark & - & - & \checkmark &- &\checkmark & \checkmark\\
{\footnotesize Autonomous NPC} &\checkmark & - & - & - &- &- & -\\
{\footnotesize Elevator}&\checkmark&-&-&-&- &- & -\\ 
{\footnotesize Long-term}&\checkmark& - & - & - & - & - & -\\
{\footnotesize Human-centered}&\checkmark& - & - & - & - & - & -\\
{\footnotesize Multi-floor} &\checkmark& - & - & - &\checkmark & - & -\\
\bottomrule 
\end{tabular}
\end{table}

\section{Dataset}
\label{Dataset}

\begin{figure}[h]
  \centering
  \includegraphics[width=0.8\linewidth]{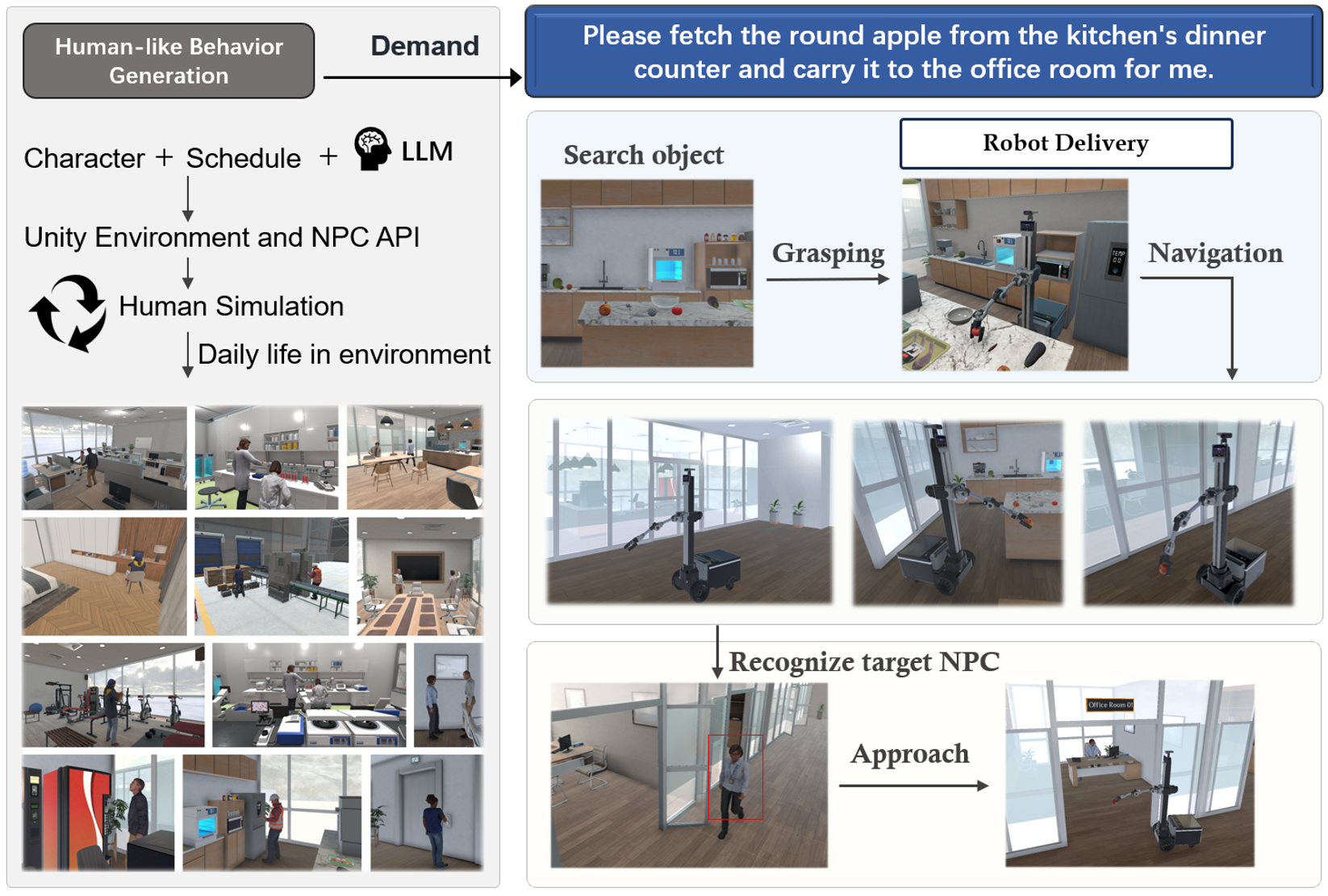}
  \caption{A data generation instance. We generate human activities, target objects, robot positions, task instructions, and a complete process of robot execution based on the settings combined with large models.}
  \label{task}
\end{figure}

\begin{table}
  \caption{The validation and test sets of the dataset encompass distinct NPC behaviors, task contexts, and linguistic instructions. We validate against potential common scenarios as a benchmark for solutions without emphasizing training and fine-tuning. However, the validation set contains appropriately annotated information, which can also be used as a training set.}
  \label{table2-dataset}
  \centering
  \begin{tabular}{lcc}
    \toprule
          & \textbf{Test Set}     & \textbf{Validation Set} \\
    \midrule
    Robot Delivery Task & 918  & 5730      \\
    Language Instructions     & 1836 & 11460    \\
    Directive Annotation Method      & GPT-4 \& GLM-4      & GLM-4  \\
    Check Method        &      Manual \& GLM-4 & GLM-4 \\
    Ground Truth Annotation  &      - & \textcolor{red}{\textbf{$\checkmark$}} \\
    Scenes       &   19  & 24 \\
    Delivery Object Categories       &    42 & 43 \\
    \bottomrule
  \end{tabular}
\end{table}


We elaborate on the data collection process, encompassing the generation of language instructions, item placement, and scene construction. Additionally, we present the data annotation methodology and results. Tasks, environments, and agents all adhere to the PRS environment settings. The task scenes are located within a three-story building in the PRS environment. Target objects encompass all interactive items, while functional equipment follows physical engines and basic logic. Language instructions originate from a task generator, refined and reviewed manually through an LLM to ensure accuracy and diversity.
Although NPCs can move and act independently in the environment \cite{li2023controllable}, and numerous interactive items and devices are present, we can configure and access all their states from the ground up, such as naming identifiers, spatial coordinates, and physical attributes. Based on the comprehensive environmental data obtained, we can generate tasks in real-time using preset scalable templates \cite{liu2019experience} and optimize task instructions with a large language model. Moreover, the environment's data interface can easily acquire relevant task information to evaluate the computation methods for generated task results.
(1) We constructed an environment-related PRS corpus, collecting verbs, nouns, and adverbs corresponding to actions, items, and locations. This corpus includes manually designed scalable templates that map verbs, nouns, and adverbs into sentences related to the current environment using reasonable grammatical rules \cite{xu2022task}. (2) We continuously generate task statements in the running simulated environment, refine and optimize them using an LLM, and finally screen them to obtain 13296 language instructions, as shown in Table \ref{table2-dataset}.

Unlike household tasks in Table \ref{table-compare}, we specialize in the indoor environment of buildings and pay particular attention to how robots can deliver items across different floors and rooms in the building based on the needs of service recipients.

\section{Baseline Method}
\label{Baseline}

\begin{figure}[h]
  \centering
  \includegraphics[width=\linewidth]{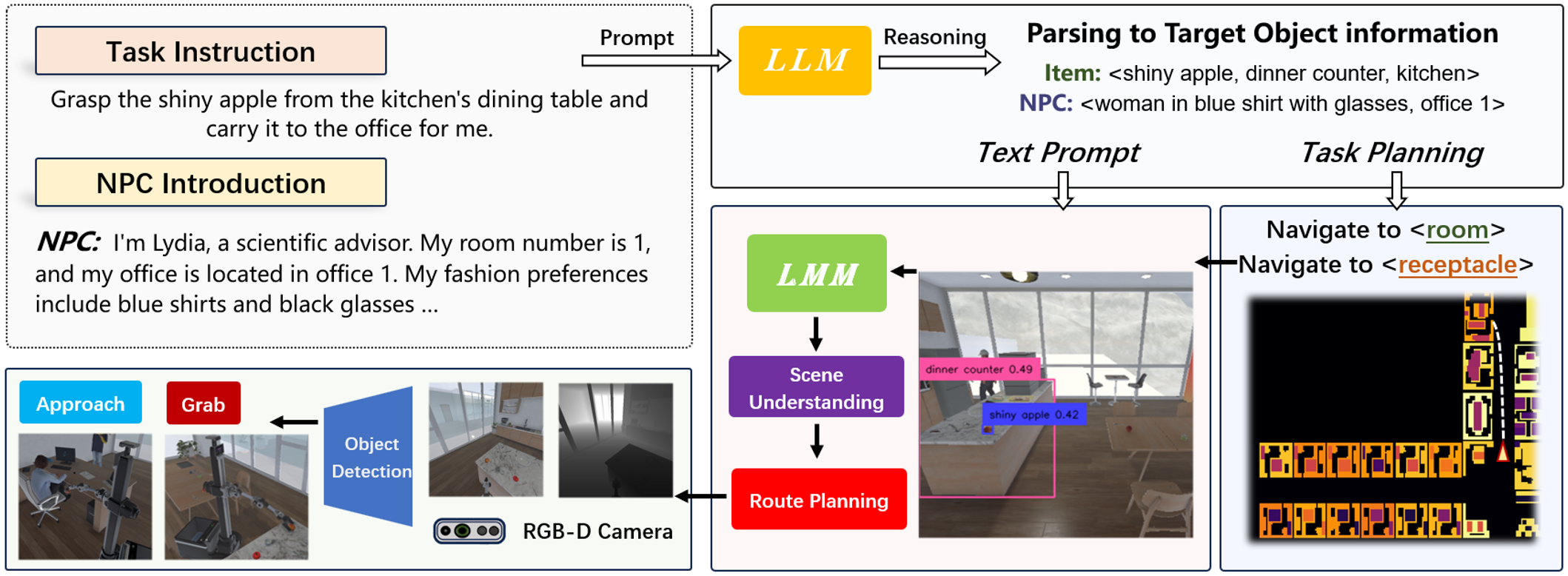}
  \caption{Modular method for the robot delivery task with LLM and LMM.}
  \label{method}
\end{figure}

The baseline method comprises multiple modules \cite{min2021film}, including the language, vision, and action modules, as shown in Figure \ref{method}, for tasks such as language parsing, navigation search, scene understanding, object recognition, segmentation, action, localization, and object manipulation.

\subsection{Language Module}The language module utilizes a large language model (LLM) to process instructions $ins$ and character introductions $intro$, outputting executable sequences based on pre-defined prompts \cite{wei2022chain}, which specify the extraction of target information and visual feature from the task context, $LLM(ins, intro, prompt)=res$. The prompt defines the output format with the fixed symbols and includes result examples to facilitate the alignment of LLM output \cite{zamfirescu2023johnny}. For instance, using the regular expression to decode relevant information, $RE(res)=<obj, recep_{obj}, room_{obj}, npc, room_{npc}>$. The executable sequences generated by LLM break down the task into subtasks, including corresponding information, such as target object search $[obj, ecep_{obj}, room_{obj}]$ (e.g., "\textit{white cup, dinner counter, kitchen}"), grasping $[obj]$, delivery to $[room_{npc}]$ (e.g., "\textit{office}"), and person search $[npc]$ (e.g., "\textit{a man with grey coat}"). Subsequently, the robot sequentially performs these subtasks to accomplish the overall task accurately.

\subsection{Vision Module}
With information (\textit{"a white cup"}) from the language module, the robot localize the specified object. The intricate spatial layout within indoor environments results in diverse positional arrangements of interactive objects, posing challenges for visual models in object detection. Furthermore, identifying diminutive, occluded, or container-enclosed objects presents a formidable obstacle \cite{zhu2021soon}.
To enhance search and recognition efficiency, we incorporate large receptacle information (\textit{"dinner counter"}) derived from task instructions, or \textit{"the apple on the dining table"}, \textit{"the water cup on the desk."} Nonetheless, conducting direct object recognition within a room for item retrieval may yield considerable errors \cite{inoue2022prompter}. 
Suppose the confidence threshold is set excessively high. In that case, the visual model may struggle to identify the target object due to the item's unknown specific location and the potential distance of the robot from it. This can result in a limited number of pixels occupied by the object within the field of view, making it challenging to distinguish its visual features. Conversely, setting the confidence threshold too low can readily lead to erroneous identifications, such as mistakenly identifying objects with similar features as the intended target.
Therefore, upon searching the target object in a room, the robot performs scene understanding using a large multimodal model (LMM) \cite{lu2023unified}. Utilizing the current image captured by the robot's camera and the textual description of the target object as input, coupled with a constraining prompt such as \textit{"Are there a white cup or large dining counter in the picture? Please answer 'yes' if so, and 'no' if not."}  The robot rotates its head camera to observe the environment and employs the output of the LMM to ascertain the target object's presence within its field of view. Once the detection of the target object is detected by the LMM, the robot proceeds to either approach the object or utilize a visual model \cite{kirillov2023segment} to recognize and segment its mask. Leveraging the robust text-image alignment capabilities of LMM allows the robot to make reasonable decisions while mitigating ineffective visual processing outcomes. The visual module capitalizes on the LMM's capacity to align comprehensive visual data with object descriptions. Conversely, the small-scale visual model focuses on local information to accurately segment the mask of the target object, thereby facilitating an efficient and precise search process.
\subsection{Action Module}
Robots rely on environmental information and task context to execute navigation, approach, and grasping actions. Following the spatial directives provided in the output of the language module, the robot navigates to the designated area, such as "the kitchen." By aligning the target room information extracted from the instructions with the scene on the semantic map, the robot uses the A-star algorithm to plan an approximate route from its current location to the target room and navigates accordingly \cite{murray2022following}.
With the segmented mask of the target object through the search strategy and recognition process, in conjunction with depth data captured by the depth camera, the robot calculates the approximate distance between itself and the target object. Should the distance exceed the desired range (1m-3m), the robot devises a local route based on map data to maneuver close to the target object. It adjusts its position and orientation by aligning its camera and body with the target object (item or NPC).
The robot accurately identifies the item and acquires its 2D mask for object grasping, serving as input parameters for the PRS grasping API to complete the target grasping action \cite{rtbpython}.

\section{Experiments}
\label{Experiments}
\subsection{Evaluation Metrics}
Two conditions determine the success of a delivery task: 1) successfully grasping the target object, which requires providing an accurate mask of the target object within an appropriate range, and 2) locating the target character, which is considered successful if the robot is within a 3D Cartesian distance of 3m from the character. The ultimate goal is to deliver the object to the vicinity of the target character. Therefore, the delivery task can be decomposed into two sub-tasks, and the experiment will evaluate the efficiency of completing these sub-tasks.
\subsection{Experimental Setup}
Task success is fulfilling all the prescribed success conditions in the instructions. During task execution, the execution results are checked after all agent robots complete their actions. Each agent is allowed only one attempt and cannot repeat the execution. Any incorrect use of interface parameters, collisions with obstacles, interactions with the wrong target, or dangerous movements will result in task failure. The task execution time is limited to 8 minutes. Access to underlying environmental data is not permitted, but all robot interfaces provided by the PRS simulator are available.
The PRS simulator offers interfaces for robot control and sensing. For the experiment, LLM employs GPT-4 and GLM-4 \cite{zeng2022glm} (given the similarity in instruction processing capabilities between GLM-4 and GLM-4V, they are not separately listed in the table), LMM utilizes GPT-4V(ison) \cite{yang2023dawn} and GLM-4V \cite{wang2023cogvlm}, and the visual detection model uses Grounding DINO \cite{liu2023grounding}, all in a zero-shot setting. The performance of each sub-module will be tested separately without other modules' results.

\subsection{Result}
With a test set of 918 tasks, the efficiency of each module was calculated in the experiment, including language parsing, object search recognition, and virtual human character search. Experimental results in Table \ref{table-experiment} were compared, and the GPT-4o-based method achieved a task success rate of 32.2\%.

\begin{table}[t]
\centering
\caption{Results on test set. \textbf{Method} includes various models for the baseline method, and GD is Grounding DINO\cite{liu2023grounding} object detection model. \textbf{Task SR} is the success rate (SR) of the complete robot delivery task, \textbf{Parsing} is the SR of language instruction parsing to correct target information, \textbf{Manipulation} represents object grasping SR, \textbf{Human Search} is human character search SR, and \textbf{Time Spent} is time used (minute) on successful task execution. }
\begin{tabular}{lc|cccc}
\toprule
\label{table1}
 \textbf{Method} & \textbf{Task SR} &\textbf{Parsing} & \textbf{Manipulation} & \textbf{Human Search} & \textbf{Time Spent }\\\cmidrule{2-2} 
\midrule
Rule-Based + GD & 3.4 & 25.9 & 14.7 & 21.1 &3.15 \\
GLM-4V & 18.7 & 68.3 & 45.7 & 78.6 & 4.68\\
GLM-4V + GD & 23.7 & 65.2 & 53.4 & 83.2 & 4.27\\
GPT-4V+ GD &28.5 & 69.7 &57.1 & 85.2 & 5.13\\
GPT-4O + GD & \textbf{\color{blue}32.2} & \textbf{\color{blue}71.7} & \textbf{\color{blue}59.2} & \textbf{\color{blue}88.7}  & 4.59\\    
\bottomrule
\end{tabular}

\label{table-experiment}
\end{table}

\section{Conclusions}

This work integrates previous work on skill-learning scenarios and explores a specific commercial scenario with human-robot interaction at its core. Specifically, we have constructed a brand-new virtual environment system for human-centred in-building delivery services, including multi-level spatial buildings, diverse functional rooms, multi-role behavior systems, robot, and item systems, as well as a delivery service dataset, and a baseline system. We believe that a significant and promising direction for the future is to integrate existing skills to simulate specific, determined commercial scenarios, ultimately aiming to drive the development of community technology toward commercialization.



\bibliographystyle{plain} 
\bibliography{main}





\newpage
\appendix

\section{Benchmark}
To investigate preliminary commercial scenarios of robotic applications, we introduce the human-centered in-building embodied delivery task, aiming to deliver specified items to the vicinity of the target human character. We developed the PRS simulation environment and collected a dataset related to delivery tasks. Consequently, the delivery task benchmark encompasses a simulator, environment API, dataset, evaluation metrics, and baseline methods, as follows:
\dirtree{%
.1 ..
.2 prs-delivery.
.3 simulator. 
    .4 Linux.
    .4 Windows.
    .4 macOS.
.3 dataset.
    .4 validation set.
    .4 test set.
.3 environment information.
    .4 semantic map.
    .4 NPC information.
    .4 Robot.urdf.
    .4 environment setting.
.3 baseline Method.
    .4 LLM.
    .4 LMM.
    .4 object detection model.
    .4 main.
.3 evaluation. 
.3 API document.  
}  

\section{Simulator}
\begin{figure}[H]
  \centering
  \includegraphics[width=0.9\linewidth]{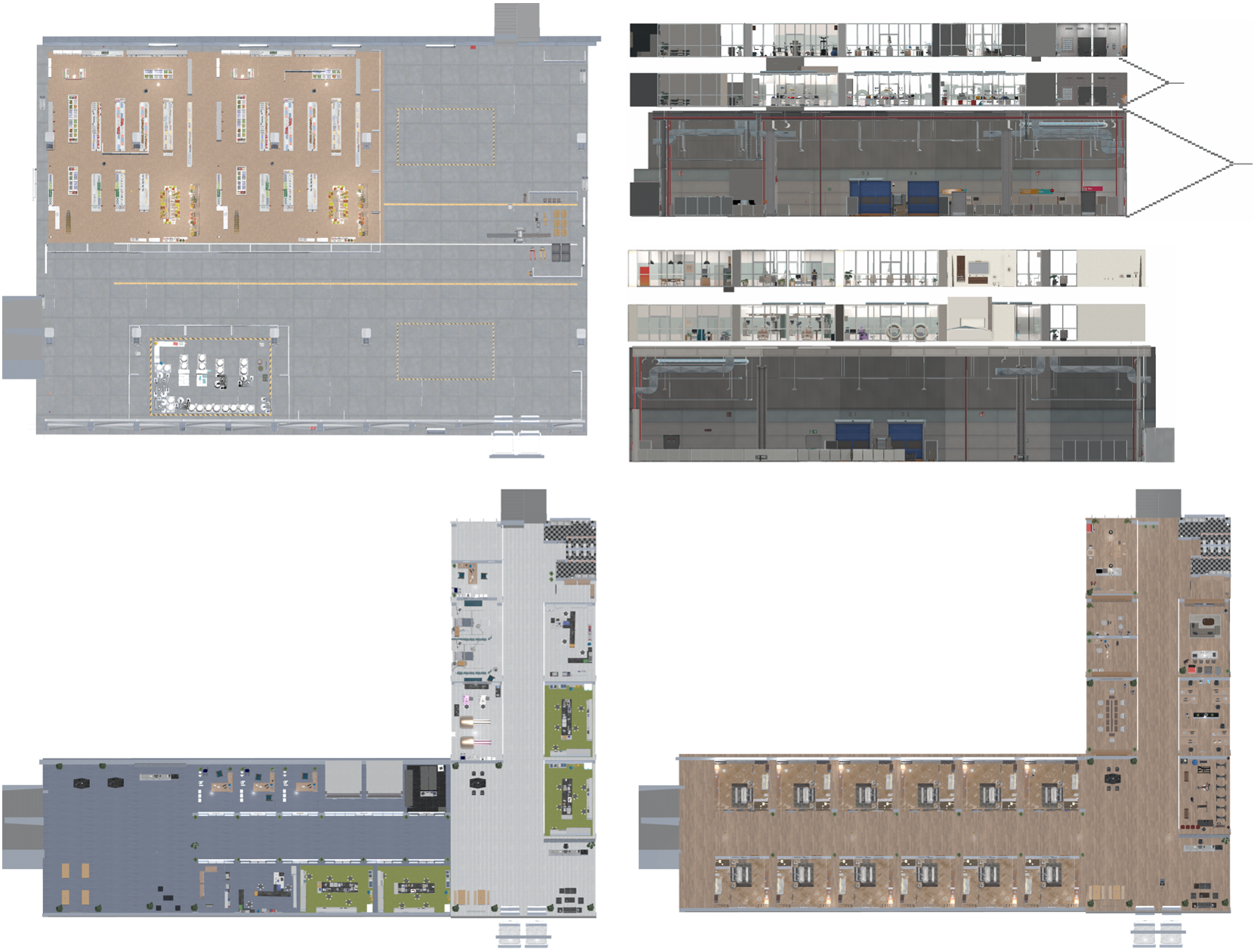}
  \caption{Two-dimensional floor plan of the polar research station building.}
  \label{app-2D}
\end{figure}

In light of our conceptualization of robotic service, and in order to better simulate comprehensive application scenarios, we have developed the Polar Research Station (PRS), a three-story building containing different rooms, providing (1) a PhysX-supported physical environment, (2) autonomous characters for performing human behaviors, (3) robots with perception sensors and interaction abilities, (4) interactive objects and devices with continuous state changes, and (5) available API for LLMs and LMMs. Figure \ref{app-2D} shows rich scenes that are close to the real world. The PRS rendering engine utilizes Unity and offers a diverse range of Python APIs. The resource is user-friendly (Python-only) and can even run without a GPU.

\section{Task Dataset}
\label{appendix-dataset}
\begin{figure}[ht]
  \centering
  \includegraphics[width=0.85\linewidth]{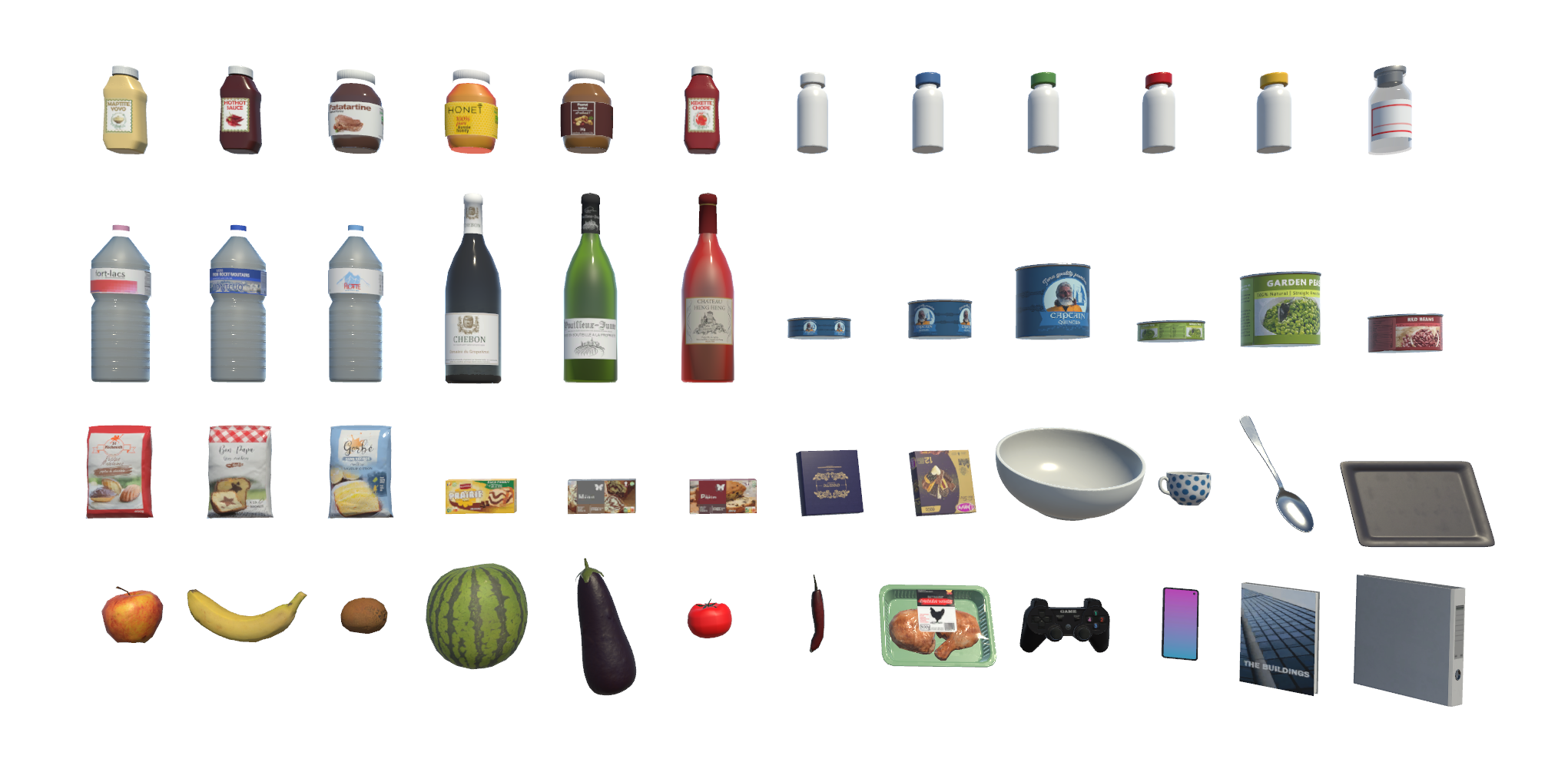}
  \vspace{-10pt}
  \caption{Interactive objects for grasping and delivery.}
  \label{app-grasp}
  
  \vspace{-10pt}
\end{figure}

\begin{figure}[h]
  \centering
  \includegraphics[width=0.85\linewidth]{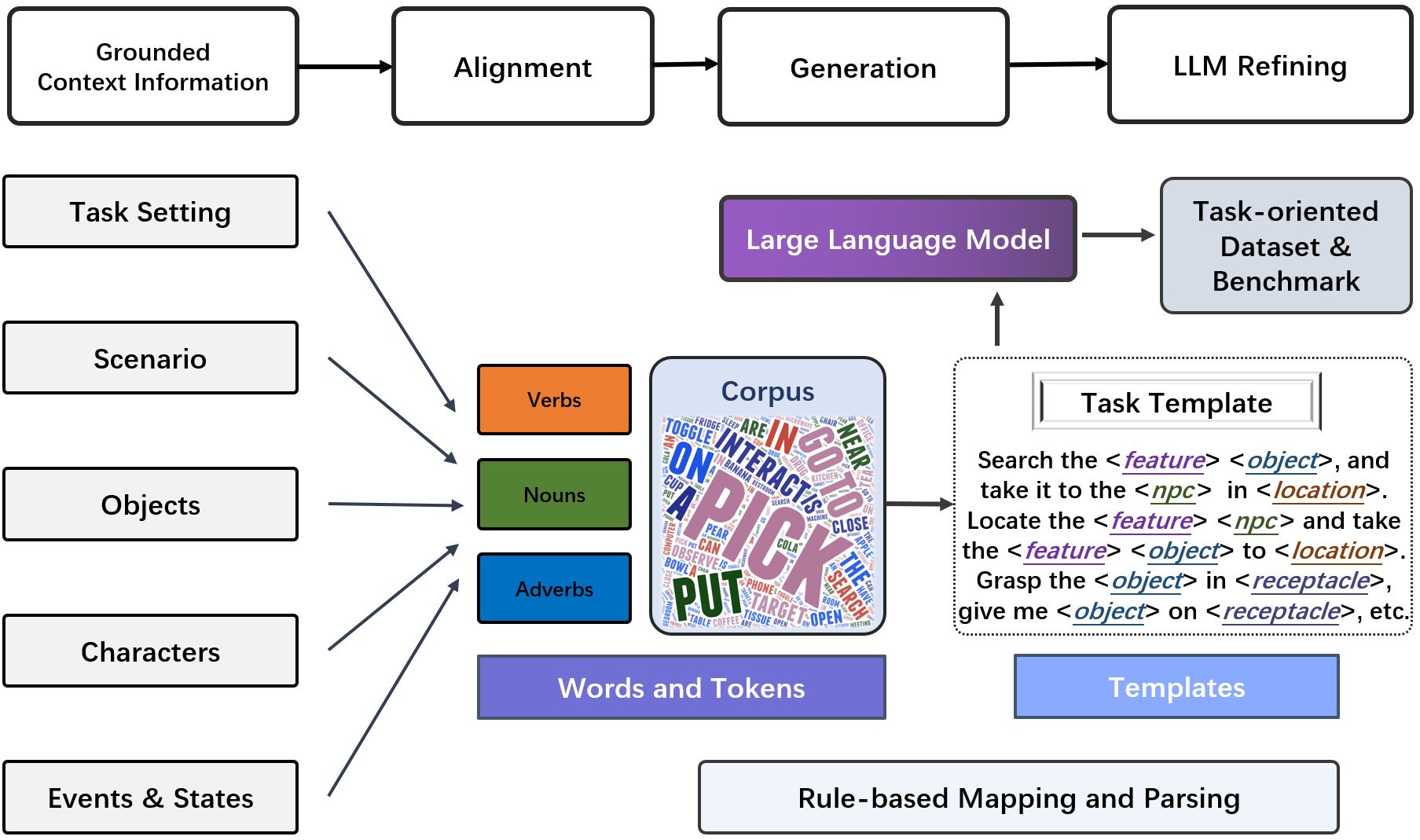}
  \caption{Automatic generation of the task instruction with template and LLM.}
  \label{app-template}
\end{figure}

\begin{figure}[h]
  \centering
  \includegraphics[width=0.75\linewidth]{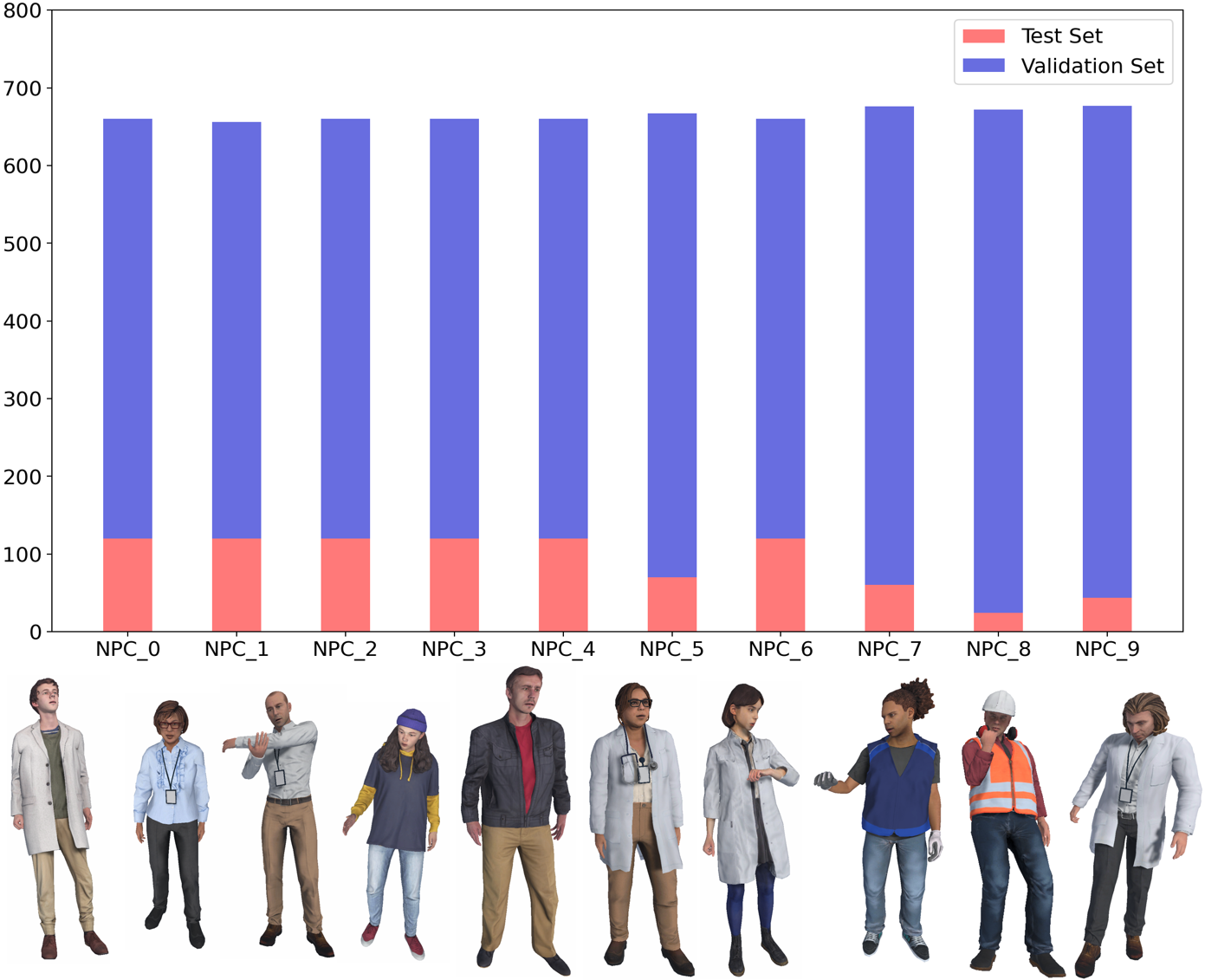}
  \vspace{-6pt}
  \caption{The frequency of NPC appearances in the dataset.}
  \label{app-npcs}
  \vspace{-10pt}
\end{figure}

\begin{figure}[h]
  \centering
  \includegraphics[width=0.98\linewidth]{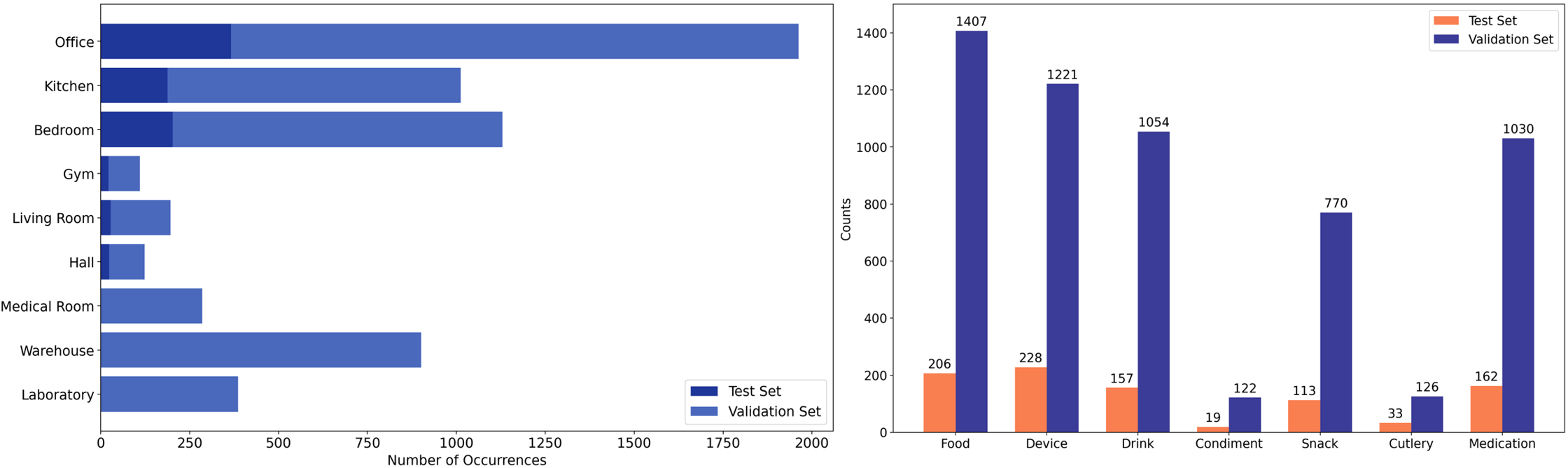}
  \vspace{-6pt}
  \caption{Statistics of different scenes and interactive objects categories in the dataset.}
  \label{app-count}
\end{figure}

The dataset is represented as a JSON file in Listing \ref{json-example}, and task initialization, execution, and evaluation are accomplished using the Python API. The task involves a variety of objects with different styles, as depicted in the Figure \ref{app-grasp}. NPCs in the environment engage in continuous simulated life activities, generating various needs over time, such as eating, drinking, working, and resting. At these moments, NPCs potentially require certain items to fulfill their demands (e.g., food, drinks, mobile phones). Thus, we simulate robot delivery services by collecting these needs. By querying environmental data, we automatically gather a large number of delivery tasks. We refine the language content using the LLM and conduct manual checks and corrections, as shown in Figure \ref{app-template}. Specifically, we introduce LMM to perform textual annotation (visual feature description) of image data to decrease manual work and increase diversity.
Figure \ref{app-npcs} indicates ten distinct NPCs as the service targets, each with their own profile and preferences. Figure \ref{app-count} illustrates the spatial distribution of scenes within the task set, demonstrating long-range visibility across spaces.

\begin{lstlisting} [label={json-example}]
{
        "task_id": "1_2025_02_11T12_45_49_10_1_1"
        "npc_name": "Imani",
        "npc_id": 1,
        "time": "2025-02-11T12:45:49",
        "npc_action": "sit",
        "npc_position": {
            "x": -16.02390480041504,
            "y": 0.0,
            "z": -8.445791244506836},
        "target_object_name": "WaterBottle_Blue_1",
        "target_object_type": "WaterBottleBlue",
        "target_object_pos": {
            "x": -16.878999710083008,
            "y": 0.7600002288818359,
            "z": -5.263000011444092},
        "directive": [
            "Grasp the blue water bottle from the wooden dining table in the kitchen and bring it to me in the kitchen room.",
            "Fetch the blue-packaged water bottle from the wooden dining table in the kitchen and deliver it to Imani, the woman in the blue shirt with black glasses, in the kitchen room."],
        "npc_description": "I'm Imani, a scientific advisor at a polar research station. My room number is 1, and my office is located in office 1. I often lead a regular life. My fashion preferences include blue shirts and black glasses."
    }
\end{lstlisting} 

\subsection{Data Augmentation}
The delivery tasks encompass 10 NPCs, 23 rooms, and 47 types of items. We utilize various NPC information (e.g., names, occupations, habits) and actions, and alter the positions of NPCs, items, and robots to enrich the benchmark distribution. For each robotic delivery task, natural language instructions with relevant context are provided to simulate the robot's instruction following.
\subsection{Dataset Split}
Unlike past supervised learning settings, we propose that embodied tasks in simulated scenarios need not be based on the independent identically distributed (IID) assumption. Consequently, we modify the setting from the traditional "train-develop-test" to "free mode-develop-test," omitting an explicit training set (with ground truth information included in the validation set for training or fine-tuning purposes). In the free mode, researchers can freely collect data without restrictions to develop and debug solutions, such as visual recognition, scene understanding, and search strategies. We argue that this setting is more advantageous for large multimodal models (zero-shot) and closer to real-world scenarios, where it is impossible to pre-acquire all user scenarios but rather to handle various potential scenarios with general solutions.
\subsection{Accessibility}
We have made the PRS simulator \footnote{\url{https://huggingface.co/datasets/xzq1999/prs-env/tree/main}} and robot delivery dataset available on \url{https://github.com/PRS-Organization/prs-delivery} and accessible to all. The simulator is provided in Linux (Ubuntu), macOS, and Windows, with continuous updates and maintenance. We explicitly offer a usable API and usage examples. Additionally, we have opened an online result evaluation \footnote{\url{https://eval.ai/web/challenges/challenge-page/2313/overview}} for the validation and test sets by Eval AI.

\subsection{Responsibility}
We are responsible for the content of the simulator and dataset, ensuring no infringement or privacy breaches. Researchers must agree to our basic terms before usage, which include taking responsibility for outcomes resulting from utilizing these resources for development, deployment, and research. We encourage researchers to open-source their code to facilitate community efforts.

\section{Delivery Process}
\begin{figure}[h]
  \centering
  \includegraphics[width=0.99\linewidth]{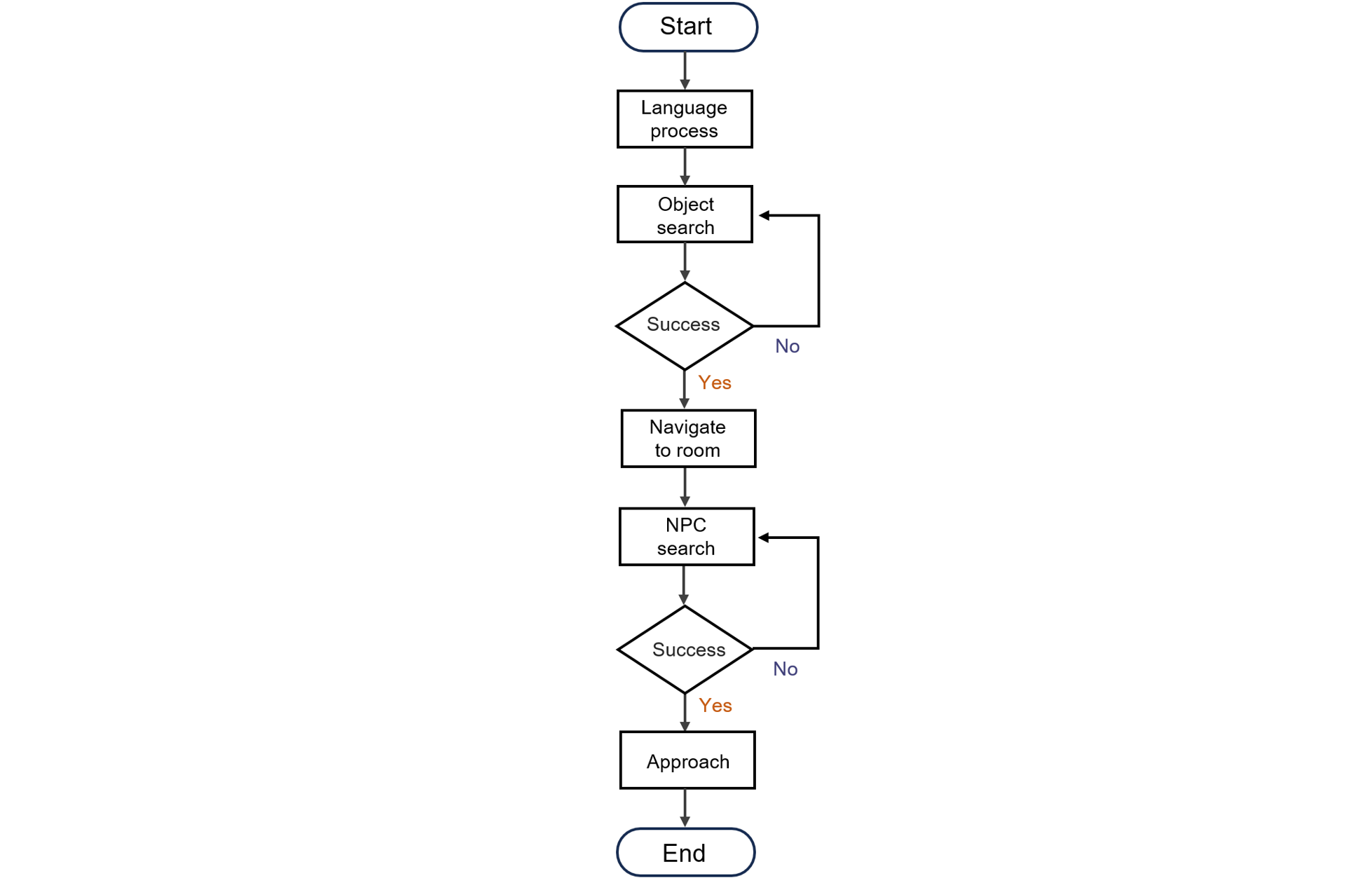}
  \caption{Flowchart of delivery task.}
  \label{app-process}
\end{figure}

\begin{figure}[h]
  \centering
  \includegraphics[width=0.99\linewidth]{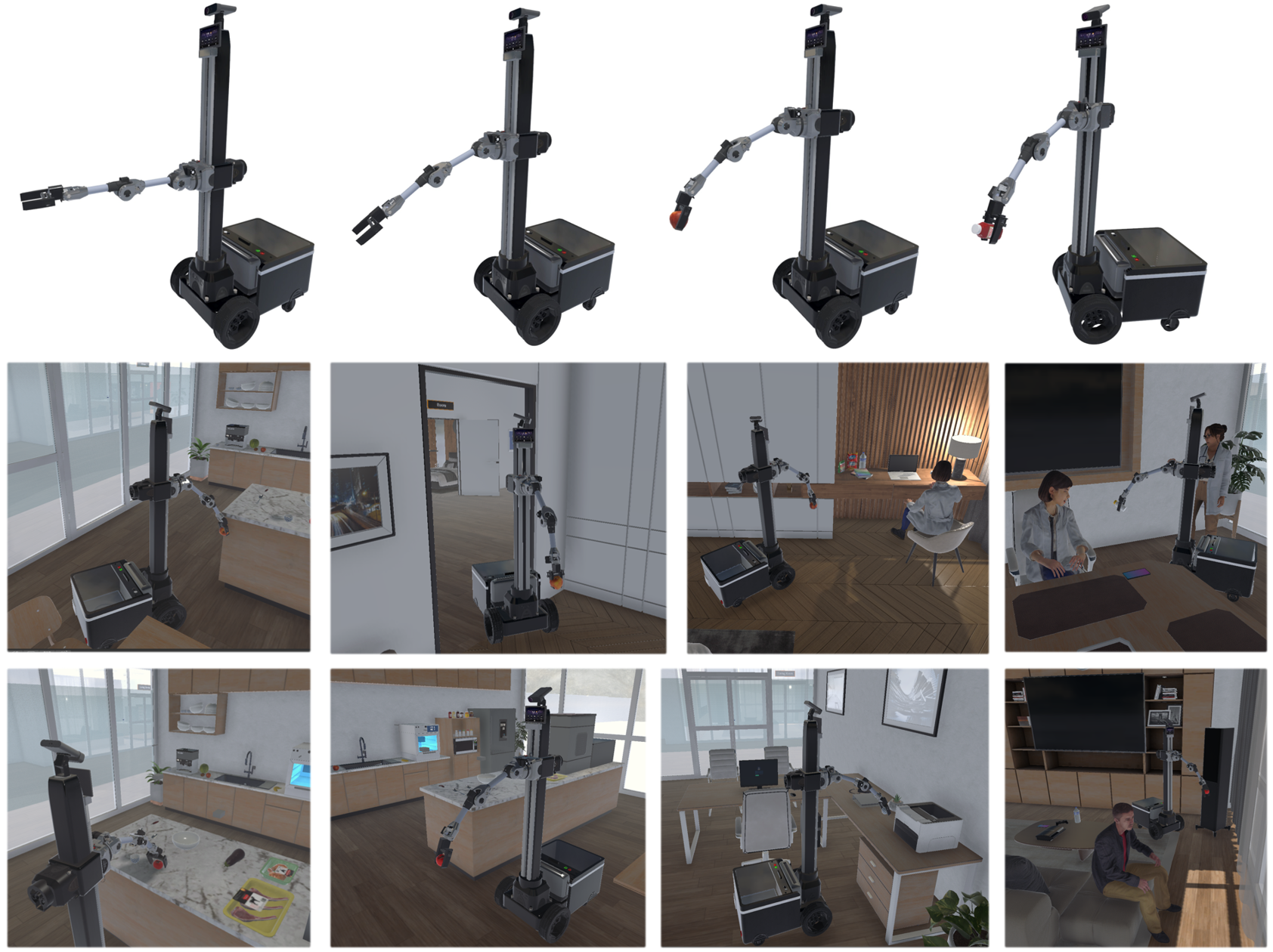}
  \caption{Delivery task examples.}
  \label{app-delivery}
\end{figure}

As shown in Figure \ref{app-process}, the task can be decomposed into several subtasks, each with explicit goals and termination conditions. The robot delivers items amidst dynamic environmental changes and NPC behaviors. Additionally, the environment features numerous interactive objects, resulting in unpredictable circumstances throughout the task, as illustrated in Figure \ref{app-delivery}.
\subsection{Robotic Skill}
In the delivery task, some executions are simplified for industrial and standard processes. Robot manipulation and navigation remain significant challenges, with different AI models addressing various robot types. We focus on comprehensive simulation of scenarios and performance evaluation without delving into the details of robot skill learning. Consequently, based on robotics standards, we provide a high-level API (ROS-like, e.g., \textit{prs.agent.goto\_target\_goal((-2.25, 0.1, -7.25), radius=1.7), prs.agent.object\_interaction(input\_matrix=segment\_matrix, manipulation=1, camera\_type=0)}) for navigation and grasping. Specifically, we offer a rough obstacle map and semantic and observation image sampling (facilitating scene comprehension and room differentiation), obviating the need for robot SLAM in large spaces during each task execution. For the robot to successfully grasp the target object, a correct segment mask must be provided within a 1.2m range, with the PRS environment already offering built-in coordinate transformation, inverse kinematics (IK) calculations, and joint control. Thus, the task solutions utilize ROS-like APIs, which abstract the specific robot model and align more closely with general algorithms. The ROS-like API setup allows for robot morphology and structure modifications at a low cost, enhancing sim2real performance.
\subsection{Baseline Reproducibility}
In the baseline method, we employ a zero-shot setting (LLM, LMM, zero-shot object detection model) instead of model fine-tuning. We have released the baseline and dataset document. This setup holds significant advantages in reproduction and secondary development. Besides replacing models, researchers can explore better prompts, target searching, navigation strategies, semantic alignment, context processing, etc., to enhance the efficiency of the robot.

\section{Limitations}
\label{appendix-limitation}
Although we have considered data augmentation and variations in style, we only constructed a three-story building and thus cannot cover all scenarios. Our dataset content has been manually verified, but the generated content of LLM and LMM may still exhibit bias and imbalance. To reduce computational expense, we simplified NPC behaviors. We simulated a robot application scenario, but the real world is far more complex and unpredictable. LLM prompts include different tone and content requirements to synthesize diverse and universal data, albeit limited to English content.

\section{Future Work}

In-building delivery is a realistic commercial scenario, differing from the popular factory assembly line scenario in that it involves more consideration of human-robot interaction. Therefore, in our future work, we will introduce (1) richer user interaction behaviors, such as users being able to send real-time location hints to the robot, and (2) longer-term user behavioral data, enabling the robot to summarize user behavior patterns for more precise service autonomously. (3) More diverse scenarios, items, and tasks. Our business scenario design, virtual environment setup, and dataset collection will iterate and continuously improve alongside research efforts in the community, commercial developments, the robotics industry, and user research.

\end{document}